\documentclass[11pt]{article}

\usepackage[preprint]{acl}
\usepackage{times}
\usepackage{latexsym}
\usepackage[T1]{fontenc}
\usepackage[utf8]{inputenc}
\usepackage{microtype}
\usepackage{inconsolata}
\usepackage{graphicx}
\usepackage{hyperref}
\usepackage{url}
\usepackage{booktabs}
\usepackage{amsfonts}
\usepackage{amssymb}
\usepackage{amsmath}
\usepackage{nicefrac}
\usepackage{microtype}
\usepackage{xcolor}
\usepackage{graphicx}
\usepackage{multirow}
\usepackage{enumitem}
\usepackage{multirow}
\usepackage{caption}
\usepackage{subfig}
\usepackage{float}
\usepackage{makecell}
\usepackage{array}
\usepackage{setspace}
\usepackage[mathlines]{lineno}
\usepackage[table]{xcolor}

\title{DeepLatent: Think with Images via Parallel Latent Visual Reasoning}

\author{
  \textbf{Dongchen Lu\textsuperscript{1*}} \hspace{1cm}
  \textbf{Zhimo Li\textsuperscript{2*}} \hspace{1cm}
  \textbf{Mao Shu\textsuperscript{1}} \hspace{1cm}
  \textbf{Huo Cao\textsuperscript{1\dag}}
\\
\\
  \textsuperscript{1}Baidu Inc.\quad
  \textsuperscript{2}Peking University
\\
  \small{* Equal contribution\quad \dag Corresponding author}
}

\begin{document}
\maketitle

\begin{abstract}
The emerging paradigm of ``thinking with images'' embeds visual states into intermediate reasoning steps, defining a new frontier for Vision-Language Models. Existing approaches diverge along two lines. Tool-assisted methods apply explicit visual operations but suffer from high latency and restricted manipulation types. Latent reasoning methods autoregressively produce implicit visual states, but underperform tool-assisted methods, and their latent tokens fail to capture effective visual information. In this work, we propose \textbf{DeepLatent}, a parallel framework for latent visual reasoning. First, we introduce LatentFormer. It uses learnable 2D tokens to generate context-conditioned latent states in parallel, anchoring every visual update directly in the original image features. Second, we design a continuous-space reinforcement learning algorithm. It optimizes latent modulation parameters directly in the embedding space, significantly improving latent representation quality. The framework is trained via knowledge distillation followed by this continuous-space RL algorithm. Furthermore, we contribute DeepLatent-180K, a large-scale dataset tailored for latent visual reasoning. Extensive evaluations across multiple benchmarks demonstrate that DeepLatent achieves state-of-the-art performance. Code is available at \url{https://github.com/ludc506/DeepLatent}.
\end{abstract}

\section{Introduction}

Recent advances in Vision-Language Models (VLMs) have enabled increasingly sophisticated visual reasoning, particularly through Chain-of-Thought (CoT) approaches \cite{llava_cot, vision_r1}. However, in mainstream VLMs \cite{llavaov, internvl3}, the reasoning trajectory is typically expressed purely in language. The visual representation remains static after a single initial encoding step. Consequently, the model cannot dynamically revisit or adapt its visual focus as reasoning unfolds. In contrast, human cognition naturally interleaves vision and reasoning, iteratively updating visual states based on an evolving understanding of the task. Bridging this gap requires embedding active visual interaction directly into the reasoning process, enabling models to genuinely ``think with images.''

\begin{figure*}[t]
    \centering
    \setlength{\abovecaptionskip}{3pt}
    \setlength{\belowcaptionskip}{-13pt}
    \includegraphics[width=\linewidth]{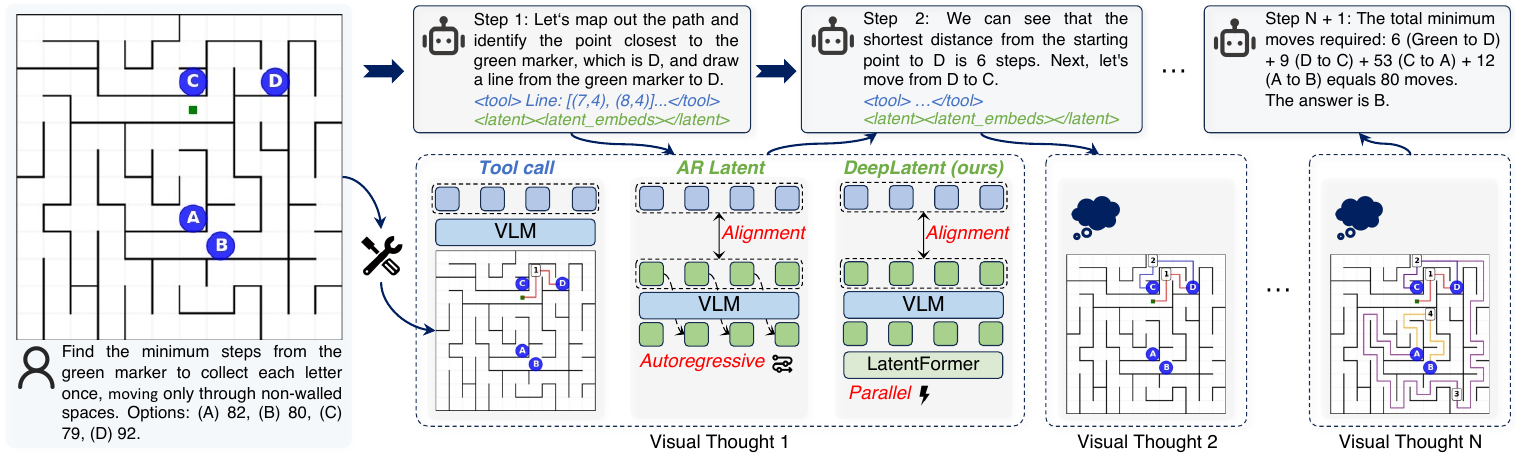}
    \caption{Interleaved visual-textual reasoning and architectural comparison. Unlike prior approaches that rely on discrete tool calls or serial autoregressive latent generation, DeepLatent employs a parallel LatentFormer to enable efficient and synchronized visual-state updates during complex reasoning.}
    \label{fig:vlm_compare}
\end{figure*}

Existing efforts mainly explore this direction in two ways. Tool-based methods manipulate images through external operations or executable code~\cite{openai_o3, deepeyes, vtoolr1, thyme}. These methods are effective and provide interpretable visual evidence, but they are limited to predefined manipulation types and often incur high inference latency. Latent visual reasoning methods take a more integrated path. 
They generate latent visual tokens in an autoregressive (AR) paradigm, internalizing intermediate visual reasoning as a model capability.
However, latent approaches still underperform tool-based methods on challenging benchmarks (e.g., a gap of up to 5\% on HRBench), and they usually require complex latent alignment.

Beyond these practical limitations, latent reasoning faces a deeper challenge. Empirical analysis~\cite{capimagine} reveals that latent tokens in previous work exhibit high inter-token homogeneity and contain limited visual information. We relate this phenomenon to two systematic biases in the AR paradigm. 
(1) \textit{Structural mismatch.} Images are continuous 2D fields whose semantics arise from spatial neighborhoods, not from a unidirectional causal chain. Forcing AR generation over flattened 1D sequences conflicts with this structure, making it difficult for latents to capture spatial semantics.  
(2) \textit{Information source deviation.} Latent visual reasoning aims to derive new visual representations from the input image. However, in AR generation, later latents are conditioned primarily on earlier latent hidden states rather than image features. This shifts the information flow from image-to-latent to latent-to-latent. 
Moreover, these latent hidden states replace the visual features typically generated by the visual encoder.
Their distributions can differ substantially from those of the encoder outputs, which may weaken the pretrained vision-language alignment.

To address these biases, we propose \textbf{DeepLatent}, a parallel, context-conditioned latent reasoning framework built on two key principles. First, it generates all latents in parallel, removing the structural mismatch of AR generation. Second, it grounds every latent directly in image features, avoiding information source deviation. At the core of DeepLatent is LatentFormer. It initializes learnable 2D latent queries and refines them through multi-layer cross-attention over image features. We further derive three modulation parameters to adapt the latents to the reasoning context. They control the latent scale, latent distribution, and image feature distribution. The generated latents are fed into the LLM in a single forward pass, enabling efficient and synchronized visual state updates. On top of this framework, we design a continuous-space reinforcement learning (RL) algorithm to optimize the modulation parameters directly in the embedding space. This further improves latent quality beyond what supervised distillation alone can achieve.

Our contributions are summarized as follows:

(1) We propose DeepLatent, a parallel and context-conditioned latent visual reasoning framework. Its LatentFormer jointly generates latent states while grounding each of them directly in visual features, improving both reasoning performance and inference efficiency.

(2) We develop a continuous-space sampling and RL algorithm for latent variable optimization. Combined with knowledge distillation, it substantially improves latent quality and yields strong performance across diverse benchmarks.

(3) We introduce DeepLatent-180K, a large-scale dataset for latent visual reasoning, which we release to facilitate further research in this domain.

\section{Related Work}

\subsection{Thinking with Images via Tools}

A line of work augments VLMs with external visual tools for dynamic image manipulation during reasoning. \textbf{Programmatic visual interfaces.} Early methods prompt LLMs to generate executable code that invokes visual APIs~\cite{vis_prog, vipergpt, mmreact}. Thyme~\cite{thyme} and related work further extend such interfaces to support richer image manipulation and distilled program execution~\cite{vpd, visualsketchpad, refocus, pyvision}. Reward-driven approaches instead learn when and how to invoke tools through RL~\cite{openthinkimg, vtoolr1}. \textbf{Visual grounding.} DyFo~\cite{dyfo} and related methods focus on task-relevant regions through bounding-box prediction or spatial search~\cite{vstar}. Some models explicit grounding steps as intermediate supervision~\cite{visual_cot, cogcom}. DeepEyes~\cite{deepeyes}, Pixel-Reasoner~\cite{pixelreasoner}, and related work~\cite{grit, mini_o3} further unify grounding and reasoning via RL. Additional efforts revisit visual evidence through explicit re-insertion or attention redistribution~\cite{qwenlookagain, lookback}.

\subsection{Latent Visual Reasoning}

Motivated by latent reasoning in LLMs~\cite{coconut, ccot}, recent work embeds visual information into continuous latents during VLM reasoning.
Monet, LVR, and Mirage~\cite{monet,lvr,mirage} supervise latent tokens via visual distillation in multi-stage pipelines, then apply RL to refine reasoning. CoVT~\cite{covt} introduces expert-specific visual tokens, aligned during training by multiple auxiliary expert heads. Others align latents to visual encoder outputs at each reasoning step~\cite{lavit,valr}, or distill visual features through teacher-guided regression~\cite{ilvr,latentsketchpad}. In contrast, some methods~\cite{mcout,mull} learn modality-agnostic tokens in a unified continuous space without explicit visual distillation. Further work explores iterative visual refocusing~\cite{lare}, memory-augmented grounding~\cite{vismem}, single-token compression~\cite{onelatent}, and predictive latent trajectory learning~\cite{pearl}. 
Despite their varied designs, these methods typically produce latent states autoregressively.

Tool-assisted methods incur high latency and limited flexibility. Latent methods avoid explicit tool overhead, but remain less competitive and retain limited visual information~\cite{capimagine}. DeepLatent bridges this gap via parallel latent generation anchored to the original image features.

\section{DeepLatent}

\begin{figure*}[t]
    \centering
    \setlength{\abovecaptionskip}{4pt}
    \setlength{\belowcaptionskip}{-16pt}
    \includegraphics[width=\linewidth]{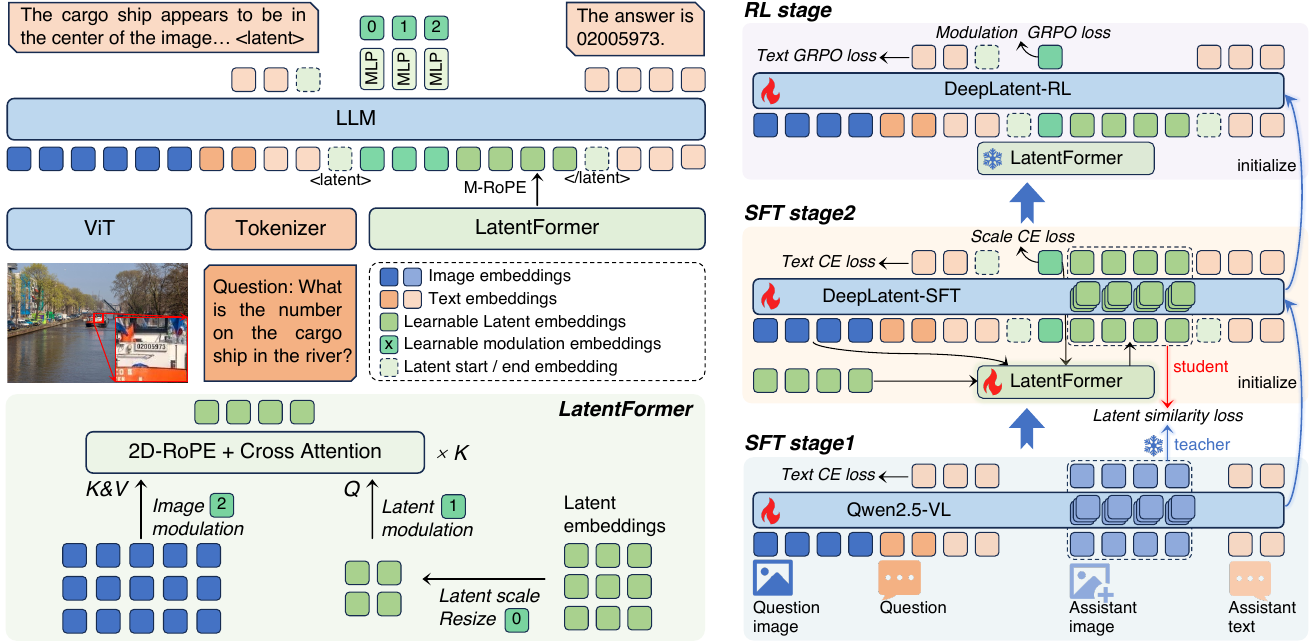}
    \caption{
    DeepLatent architecture and training pipeline. Left: At each \texttt{<latent>} position, the LLM modulates the LatentFormer to generate latents and insert them into the sequence. Right: SFT Stage 1 trains Qwen2.5-VL on interleaved image-text data and caches assistant image hidden states. SFT Stage 2 trains the LatentFormer to reconstruct cached visual states. RL stage jointly optimizes text generation and latent modulation via GRPO.
    }
    \label{fig:framework}
\end{figure*}

As illustrated in Fig.~\ref{fig:framework}, DeepLatent employs a dedicated LatentFormer for parallel latent generation, bypassing autoregressive latent prediction inside the LLM.
When the VLM emits a \texttt{<latent>} token, a set of learnable 2D latent embeddings interacts with the question image features via cross-attention, forming a spatial latent grid grounded in the image.
These latents are inserted jointly into the reasoning sequence as internal visual states, so the LLM can process the full grid in a single forward pass rather than one token at a time.

Built on Qwen2.5-VL~\cite{qwen25vl}, DeepLatent is trained with a three-stage pipeline, following prior latent visual reasoning methods. In SFT Stage 1, the model is trained on image-text interleaved data with assistant images, and the corresponding LLM hidden states are cached as teacher visual states. In SFT Stage 2, the assistant images are removed, and the LatentFormer is trained to reconstruct the cached teacher states with a cosine similarity loss, together with next-token prediction and latent scale losses. In the RL stage, we apply GRPO~\cite{grpo} to further refine latent invocation and optimize latent modulation parameters in continuous space.

\subsection{SFT Stage 1: Interleaved Reasoning}

Image-text interleaved reasoning trajectories provide rich supervision, as they include assistant images that serve as intermediate visual context for subsequent textual reasoning. In SFT Stage 1, we train the model with standard next-token prediction over complete trajectories:
\begingroup
\setlength{\abovedisplayskip}{4pt} 
\setlength{\belowdisplayskip}{4pt}
\begin{equation}
    \mathcal{L}_{\mathrm{S1}} = -\!\sum_{t \in \mathcal{T}} \log p_\theta(y_t \mid y_{<t}, q, I_0, I_{1:N}),
\end{equation}
\endgroup
where $y_t$ is the ground-truth response token at position $t \in \mathcal{T}$, $q$ and $I_0$ are the input question text and image, and $I_{1:N}$ are the assistant images. After training, we perform a forward pass over the same trajectories and cache the LLM hidden states corresponding to each assistant image. 
These cached ``teacher visual states'' will supervise the latent distillation in SFT Stage 2.

\subsection{SFT Stage 2: Latent Distillation}

In SFT Stage 2, explicit assistant images in the training trajectories are replaced by latent blocks. DeepLatent learns to recover the visual states previously provided by these images, conditioned on both the reasoning context and the question image.

\paragraph{Latent Construction.}
When the model emits a \texttt{<latent>} token, we interrupt the autoregressive decoding and inject three learnable embeddings: a scale-control embedding $\mathbf{e}_s$, a latent-modulation embedding $\mathbf{e}_l$, and an image-modulation embedding $\mathbf{e}_i$. The LLM then produces corresponding hidden states $\mathbf{h}_s$, $\mathbf{h}_l$, and $\mathbf{h}_i$ conditioned on the current reasoning context. 
Their functional roles are strictly decoupled: $\mathbf{h}_s$ dictates the latent spatial scale; $\mathbf{h}_l$ modulates the queries to drive visual search (e.g., localizing relevant regions); and $\mathbf{h}_i$ modulates the question image features to perform implicit visual editing (e.g., synthesizing auxiliary structures). The LatentFormer constructs the latent embeddings $\mathbf{Z}$ from these states, after which a \texttt{</latent>} token closes the block and text generation resumes. The complete latent sequence takes the form: \texttt{<latent>}$\,\mathbf{e}_s\;\mathbf{e}_l\;\mathbf{e}_i\;\mathbf{Z}\,$\texttt{</latent>}.

\textbf{Spatial latent queries.}
We maintain a learnable latent grid $\mathbf{L} \in \mathbb{R}^{M_H \times M_W \times d}$ as the query template for LatentFormer, where $(M_H, M_W)$ denotes the maximum grid size. A latent scale head takes $\mathbf{h}_s$ as input and predicts the grid height and width as two independent discrete distributions:
\begingroup
\setlength{\abovedisplayskip}{4pt}
\setlength{\belowdisplayskip}{4pt}
\begin{equation}
\begin{aligned}
\hat{\mathbf{p}}_H &= \mathrm{MLP}_h(\mathbf{h}_s), \quad \hat{H}_z = \mathop{\arg\max}(\hat{\mathbf{p}}_H), \\
\hat{\mathbf{p}}_W &= \mathrm{MLP}_w(\mathbf{h}_s), \quad \hat{W}_z = \mathop{\arg\max}(\hat{\mathbf{p}}_W).
\end{aligned}
\end{equation}
\endgroup
To decouple scale learning from latent construction, we use the ground-truth (GT) grid size $(H_z^*, W_z^*)$, derived from the assistant image feature dimensions, for latent generation during training, and use the predicted size $(\hat{H}_z, \hat{W}_z)$ at inference. The query grid $\mathbf{L}$ is then spatially resized to $H_z \times W_z \times d$.

\textbf{Stochastic affine modulation.}
For each modulation state $\mathbf{h}_x$ ($x \in \{l, i\}$), an independent MLP predicts channel-wise modulation parameters $[\boldsymbol{\gamma}_x, \boldsymbol{\beta}_x, \log \boldsymbol{\sigma}_x] \in \mathbb{R}^{1 \times 1 \times 3d}$, 
which are broadcast across spatial dimensions. We then modulate the resized latent queries $\mathbf{L}$ and the question image features $\mathbf{I}_0 \in \mathbb{R}^{H_I \times W_I \times d}$ as: 
\begingroup
\setlength{\abovedisplayskip}{4pt}
\setlength{\belowdisplayskip}{4pt}
\begin{equation}\label{eq:modulation}
\begin{aligned}
\tilde{\mathbf{L}} &= \boldsymbol{\gamma}_l \odot \mathbf{L} + \boldsymbol{\beta}_l
+ \boldsymbol{\sigma}_l \odot \boldsymbol{\epsilon}_l, \\
\tilde{\mathbf{I}} &= \boldsymbol{\gamma}_i \odot \mathbf{I}_0 + \boldsymbol{\beta}_i
+ \boldsymbol{\sigma}_i \odot \boldsymbol{\epsilon}_i,
\end{aligned}
\end{equation}
\endgroup
where $\boldsymbol{\epsilon}_l$ and $\boldsymbol{\epsilon}_i$ are i.i.d. $\mathcal{N}(0, 1)$ noise. This operation couples a deterministic affine transformation with a reparameterized stochastic perturbation. The injected noise acts as an implicit regularizer during training, promoting a smoother latent space and better generalization. At inference, the stochastic term is omitted for deterministic execution.

\textbf{LatentFormer.}
LatentFormer is a lightweight latent transformer with $K$ stacked blocks. Starting from the initial latent grid $\mathbf{Z}^{(0)} = \tilde{\mathbf{L}}$, each block applies self-attention to refine the latent features, followed by cross-attention with the modulated image features $\tilde{\mathbf{I}}$. The $k$-th block computes:
\begingroup
\setlength{\abovedisplayskip}{2pt}
\begin{equation}
\begin{aligned}
\mathbf{Z}' &\!=\!\mathbf{Z}^{(k-1)} \!+\! \mathrm{SelfAttn} \bigl( R_{\mathrm{2D}}(\mathbf{Z}^{(k-1)}) \bigr), \\
\hspace{-0.3em}
\mathbf{Z}^{(k)} &\!=\!\mathbf{Z}' \!+\! \mathrm{CrossAttn} \bigl( R_{\mathrm{2D}}(\mathbf{Z}'),\,\! R_{\mathrm{2D}}(\tilde{\mathbf{I}}),\,\! \tilde{\mathbf{I}} \bigr),
\end{aligned}
\end{equation}
\endgroup
where $R_{\mathrm{2D}}(\cdot)$ denotes 2D RoPE~\cite{qwen2vl}. The final latent representation $\mathbf{Z} = \mathbf{Z}^{(K)} \in \mathbb{R}^{H_z \times W_z \times d}$ is flattened into $H_zW_z$ latent tokens and inserted into the designated latent positions of the LLM input sequence.

\paragraph{Loss Function.}
SFT Stage 2 objective consists of three losses. First, we employ a next-token prediction loss to supervise discrete generation:
\begingroup
\setlength{\abovedisplayskip}{4pt}
\setlength{\belowdisplayskip}{4pt}
\begin{equation}
\mathcal{L}_{\mathrm{CE}} = -\!\textstyle\sum_{t \in \mathcal{T}} \log p_\theta(y_t \mid y_{<t}, q, I_0, \mathbf{Z}_{1:N}),
\end{equation}
\endgroup
where $\mathcal{T}$ denotes the positions of discrete text and \texttt{<latent>} tokens, and the continuous latent embeddings $\mathbf{Z}_{1:N}$ serve only as conditioning context.

Second, we align the student's latent-block hidden states with the cached teacher visual states from Stage 1 via a cosine similarity loss:
\begingroup
\setlength{\abovedisplayskip}{4pt} 
\setlength{\belowdisplayskip}{4pt}
\begin{equation}
\mathcal{L}_{\mathrm{Align}} \!=\! \frac{1}{|\mathcal{D}||\mathcal{P}|} \sum_{d \in \mathcal{D}} \sum_{p \in \mathcal{P}} \bigl( 1 \!- \!\cos( \mathbf{h}^{\mathrm{T}}_{d,p}, \mathbf{h}^{\mathrm{S}}_{d,p} ) \bigr)
\end{equation}
\endgroup
where $\mathbf{h}^{\mathrm{T}}_{d,p}$ and $\mathbf{h}^{\mathrm{S}}_{d,p}$ denote the teacher and student hidden states at layer $d$ and position $p$, respectively. Here, $\mathcal{D}$ denotes the selected LLM layers used for alignment, and $\mathcal{P}$ denotes the alignment positions.

Third, we supervise the latent spatial scale:
\begingroup
\setlength{\abovedisplayskip}{4pt}
\setlength{\belowdisplayskip}{4pt}
\begin{equation}
\mathcal{L}_{\mathrm{Scale}} = \mathrm{CE}(\hat{\mathbf{p}}_H, H_z^{*}) + \mathrm{CE}(\hat{\mathbf{p}}_W, W_z^{*}),
\end{equation}
\endgroup
where $\hat{\mathbf{p}}_H\!\in\!\mathbb{R}^{M_H},\hat{\mathbf{p}}_W\!\in\!\mathbb{R}^{M_W}$ are the predicted scale logits, and $(H_z^{*}, W_z^{*})$ are the GT grid sizes. 

The final loss is formulated as:
\begingroup
\setlength{\abovedisplayskip}{4pt}
\setlength{\belowdisplayskip}{4pt}
\begin{equation}
\mathcal{L}_{\mathrm{S2}} = \mathcal{L}_{\mathrm{CE}} + \lambda_{\mathrm{Align}}\mathcal{L}_{\mathrm{Align}} + \lambda_{\mathrm{Scale}}\mathcal{L}_{\mathrm{Scale}},
\end{equation}
\endgroup
in which we set $\lambda_{\mathrm{Align}}=4.0$ and $\lambda_{\mathrm{Scale}}=0.5$.

\begin{figure*}[t]
    \centering
    \setlength{\abovecaptionskip}{5pt}
    \setlength{\belowcaptionskip}{-13pt}
    \includegraphics[width=\linewidth]{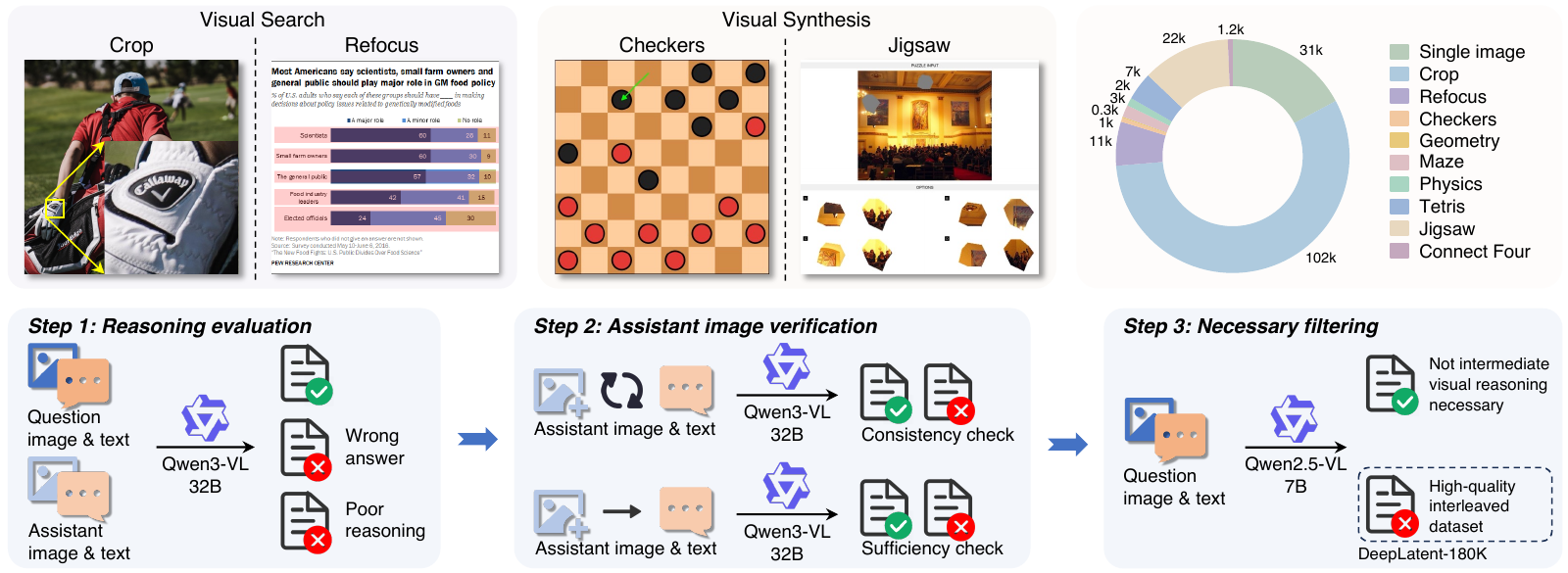}
    \small
    \caption{Overview of the DeepLatent-180K dataset composition and the automated filtering pipeline.}
    \label{fig:deeplatent_180k}
\end{figure*}

\subsection{Reinforcement Learning}\label{sec:rl}

To further refine the model's reasoning capabilities, we apply reinforcement learning from verifiable rewards via GRPO. Our objective is to jointly optimize the discrete decision of \emph{when} to invoke a latent block and the continuous decision of \emph{how} to modulate the latent embeddings. To preserve the latent construction pipeline learned during SFT, we freeze the visual encoder, LatentFormer, and $\mathrm{MLP}_s$. Only the LLM and the modulation MLPs ($\mathrm{MLP}_l$, $\mathrm{MLP}_i$) are updated in this stage.

\textbf{Continuous-space GRPO.}
Standard GRPO is defined for discrete text tokens and cannot directly optimize stochastic modulation. We therefore extend the action space to include both discrete actions (text and \texttt{<latent>} tokens) and continuous modulation actions produced by $\mathrm{MLP}_l$ and $\mathrm{MLP}_i$. The stochastic modulation in Eq.~\eqref{eq:modulation} naturally defines the corresponding action distributions. Under policy $\pi_\theta$, the modulation MLPs predict Gaussian parameters for latent query and image feature modulation. Given the sampled modulation outputs $\tilde{\mathbf{L}}^{\mathrm{old}}$ and $\tilde{\mathbf{I}}^{\mathrm{old}}$ from rollout, their log-probabilities under the current policy are computed as:
\begingroup
\setlength{\abovedisplayskip}{4pt}
\setlength{\belowdisplayskip}{4pt}
\begin{equation}
\begin{aligned}
\log\pi_\theta(\tilde{\mathbf{L}}^{\mathrm{old}}) &= \log\mathcal{N}\!\bigl(
\tilde{\mathbf{L}}^{\mathrm{old}};\; \boldsymbol{\mu}_\theta^l,\;
(\boldsymbol{\sigma}_\theta^l)^2\bigr), \\
\log\pi_\theta(\tilde{\mathbf{I}}^{\mathrm{old}}) &= \log\mathcal{N}\!\bigl(
\tilde{\mathbf{I}}^{\mathrm{old}};\; \boldsymbol{\mu}_\theta^i,\;
(\boldsymbol{\sigma}_\theta^i)^2\bigr),
\end{aligned}
\end{equation}
\endgroup
where $\boldsymbol{\mu}_\theta^l = \boldsymbol{\gamma}_{\theta,l} \odot \mathbf{L} +
\boldsymbol{\beta}_{\theta,l}$, and $\boldsymbol{\mu}_\theta^i = \boldsymbol{\gamma}_{\theta,i}
\odot \mathbf{I}_0 + \boldsymbol{\beta}_{\theta,i}$. 
We sum the log-densities over all dimensions of the sampled modulation tensor. The sampled modulations are then deterministically mapped to the latent block $\mathbf{Z}$ by the frozen latent module. Hence, $\mathbf{Z}$ itself is not treated as an RL action.

\textbf{GRPO objective.}
For each image-question pair $(I_0, q)$, the policy samples $G$ trajectories $\{\mathbf{o}_g\}_{g=1}^G$  and assigns each trajectory an outcome reward. A normalized advantage is computed as: 

\begingroup
\setlength{\abovedisplayskip}{0pt}
\setlength{\belowdisplayskip}{4pt}
\begin{equation}
A^{(g)} = \frac{R^{(g)} - \mathrm{mean}(\{R^{(j)}\}_{j=1}^{G})}
{\mathrm{std}(\{R^{(j)}\}_{j=1}^{G})}.
\end{equation}
\endgroup

We update the policy $\pi_\theta$ by maximizing:
\begingroup
\setlength{\abovedisplayskip}{4pt}
\setlength{\belowdisplayskip}{4pt}
\begin{equation}
\begin{split}
\mathcal{J}(\theta) &= \mathbb{E}_{I_0, q, \mathbf{o} \sim \pi_{\mathrm{old}}}
\frac{1}{G} \sum_{g=1}^{G} \frac{1}{|\mathcal{M}^{(g)}|}
\sum_{t \in \mathcal{M}^{(g)}} \Bigl[ \\
&\quad \min\!\bigl( r_{g,t} A^{(g)},
\operatorname{clip}(r_{g,t}, 1-\epsilon, 1+\epsilon) A^{(g)} \bigr) \\
&\quad - \beta_{\mathrm{KL}} \mathbb{D}_{\mathrm{KL}}
( \pi_{\theta} \| \pi_{\mathrm{ref}} ) \Bigr].
\end{split}
\raisetag{\baselineskip}
\end{equation}
\endgroup
Here, $\mathcal{M}^{(g)}$ indexes all policy actions in trajectory $g$, including text tokens, \texttt{<latent>} tokens, and continuous modulation actions, and $|\mathcal{M}^{(g)}|$ is the number of such actions. At step $t$, the importance sampling ratio is $r_{g,t} = \frac{\pi_\theta(\mathbf{o}_{g,t} \mid I_0, q, \mathbf{o}_{g,<t})}
{\pi_{\theta_{\mathrm{old}}}(\mathbf{o}_{g,t} \mid I_0, q, \mathbf{o}_{g,<t})}.$ Both $r_{g,t}$ and $\mathbb{D}_{\mathrm{KL}}$ take the corresponding form for the action type at step $t$: categorical probabilities and standard KL divergence for discrete tokens, and Gaussian densities and closed-form KL for continuous modulations. The reference policy $\pi_{\mathrm{ref}}$ is frozen from SFT Stage~2.

\subsection{DeepLatent-180K}\label{sec3.4}
High-quality interleaved image-text data is essential for DeepLatent-SFT, yet existing datasets exhibit two limitations: many interleaved samples can be answered without intermediate images (providing no valid supervision for intermediate visual reasoning), and current Visual Search data lacks sufficient resolution for fine-grained details. To address this, we introduce DeepLatent-180K, augmenting existing datasets with high-resolution images and rigorous filtering to ensure all samples strictly require intermediate visual reasoning.

As summarized in Fig. \ref{fig:deeplatent_180k}, the dataset comprises \textit{single-image} and \textit{interleaved reasoning} samples. The latter is categorized into two main types: \textbf{Visual Search} (e.g., cropping or refocusing on relevant regions) and \textbf{Visual Synthesis} (e.g., constructing intermediate states for mazes, jigsaw).

We construct the training data from two primary sources. First, we curate samples from existing VLM datasets (Zebra-CoT~\cite{zebra_cot}, ReFocus, CogCoM, Thyme-SFT). Second, to enhance visual perception capacity, we utilize high-resolution images from SAM-1B~\cite{sam}, FineRS~\cite{finers}, V*, InfoVQA~\cite{infovqa}, and PosterSum~\cite{postersum}. For these images, we prompt Qwen3-VL-32B~\cite{qwen3vl} to generate question-answer pairs targeting local details. These images and questions are subsequently fed into Thyme-7B to produce complete interleaved reasoning samples.

To ensure data quality and reasoning necessity, we design a three-stage automated filtering pipeline. \textbf{(1) Reasoning evaluation.} Qwen3-VL-32B evaluates the full trajectory, discarding incorrect answers or flawed logic. \textbf{(2) Assistant image verification.} The same model checks each intermediate assistant image for consistency (aligning visually with the intermediate text) and sufficiency (providing enough evidence to derive the final answer), discarding any failures. \textbf{(3) Necessity filtering.} We evaluate whether Qwen2.5-VL-7B can correctly answer the question without access to the assistant images. If correct, we discard these simple samples. Ultimately, this rigorous pipeline yields the high-quality DeepLatent-180K dataset.

\begin{table*}[t]
\centering
\setlength{\abovecaptionskip}{5pt}
\setlength{\belowcaptionskip}{-13pt}
\small
\setlength{\tabcolsep}{1.7pt} 

\begin{tabular}{l | cccccc | cccc | ccc}
\toprule
\multirow{2}{*}{Model} & \multicolumn{6}{c|}{High-Resolution Perception} & \multicolumn{4}{c|}{General Evaluation} & \multicolumn{3}{c}{Visual Reasoning} \\
 & V* & HR4K & HR8K & \begin{tabular}[c]{@{}c@{}}MME-\\ RW\end{tabular} & \begin{tabular}[c]{@{}c@{}}MME-\\ RW-Lite\end{tabular} & \begin{tabular}[c]{@{}c@{}}Visual\\ Probe-H\end{tabular} & \begin{tabular}[c]{@{}c@{}}OCR\\ Bench\end{tabular} & \begin{tabular}[c]{@{}c@{}}Chart\\ QA\end{tabular} & MMStar & RWQA & \begin{tabular}[c]{@{}c@{}}Visu\\ Logic\end{tabular} & \begin{tabular}[c]{@{}c@{}}Tree\\ Bench\end{tabular} & \begin{tabular}[c]{@{}c@{}}Baby\\ Vision\end{tabular} \\

\specialrule{\lightrulewidth}{\aboverulesep}{0pt} 
\rowcolor{blue!3} \multicolumn{14}{c}{\textit{Open-source Models}} \\ 

LLaVA-OneVision-7B      & 75.4      & 63.0      & 59.8      & 57.4      & 48.5      & 13.4      & -         & 80.0      & 61.9      & -         & -          & 37.3     & - \\
Qwen2.5-VL-7B    & 76.5      & 68.8      & 65.3      & 57.3      & 44.3      & 23.9      & 86.4      & 86.2      & 61.9      & 67.4      & 20.0       & 37.0     & 12.9 \\
InternVL3-8B      & 81.2      & 70.0      & 69.3      & 61.2      & 48.6      & -         & \textbf{88.0}      & 86.0      & \textbf{68.2}      & 70.8      & 24.9       & 38.5      & 12.9 \\

\specialrule{\lightrulewidth}{\aboverulesep}{0pt}
\rowcolor{orange!3} \multicolumn{14}{c}{\textit{Tool-based Models}} \\ 

Pixel-Reasoner-7B  & 84.3      & 74.0      & 66.9      & 64.4      & -         & 28.8      & -         & -         & -         & -         & -         & 39.0  & - \\
DeepEyes-7B       & \textbf{85.6}      & 75.1      & 72.6      & 64.1      & 53.2      & 35.1      & -         & -         & -         & -         & -         & 37.5  & - \\
Thyme-7B          & 82.2      & 77.0      & 72.0      & \textbf{64.8}      & 55.2      & -         & 86.3      & 86.1      & 65.9      & 70.2      & 23.4       & -     & - \\
DyFo-7B          & 84.3      & 71.3      & 69.8      & -         & -         & -         & -         & -         & -         & -         & -     & -     & - \\

\specialrule{\lightrulewidth}{\aboverulesep}{0pt}
\rowcolor{green!3} \multicolumn{14}{c}{\textit{Latent Visual reasoning Models}} \\ 

LVR-7B          & 81.7      & 70.8      & 63.0      & -         & 50.6      & -         & -         & -         & -         & 67.7      & -     & 39.0  & - \\
Monet-7B        & 83.3      & 71.0      & 68.0      & -         & \textbf{55.5}      & -         & -         & -         & -         & -         & -     & -     & - \\
CoVT-7B          & 78.5      & 72.5      & 69.9      & 63.3      & -         & -         & -         & -         & -         & \textbf{71.8}      & -     & -     & - \\
\midrule
DeepLatent-SFT-7B    & 84.3      & 76.4      & 73.6      & 63.1      & 52.4      & 35.0      & 85.3      & 84.2      & 63.6      & 71.0      & 24.8      & 37.9  & 14.4 \\
DeepLatent-RL-7B*    & 85.3      & \textbf{77.9}      & \textbf{74.1}      & 64.1      & 54.1      & 37.7      & 86.2      & 85.9      & 64.1      & 70.6      & 24.5      & 37.3  & 14.9 \\
DeepLatent-RL-7B     & 84.8      & 77.3      & 74.0      & 64.2      & 53.7      & \textbf{38.7}      & 86.4      & \textbf{86.4}      & 65.0      & 71.5      & \textbf{25.3}      & \textbf{39.0}  & \textbf{16.2} \\
\rowcolor{gray!5} $\Delta$ (\textit{vs} Qwen2.5-VL-7B) & +8.3 & +8.5 & +8.7 & +6.9 & +9.4 & +14.8   & +0.0 & +0.2 & +3.1 & +4.1 & +5.3 & +2.0 & +3.3 \\
\bottomrule
\end{tabular}
\caption{Performance comparison of different benchmarks.
Best results are in \textbf{bold}, and the $\Delta$ row reports absolute gains of DeepLatent-RL-7B over Qwen2.5-VL-7B.
DeepLatent* uses only visual search data of DeepLatent-180K in SFT Stage~2, while DeepLatent uses the full DeepLatent-180K dataset.}
\label{tab:evaluation_results}
\end{table*}

\section{Experiments}

\subsection{Experimental Setup}

\textbf{Training Configuration.} We build DeepLatent on top of Qwen2.5-VL-7B. In SFT Stage 1, the model is trained on the full DeepLatent-180K dataset. In SFT Stage 2, we train two variants: \textbf{DeepLatent}, which uses the full DeepLatent-180K mixture, and \textbf{DeepLatent*}, which uses only the visual search subset to emphasize high-resolution localization. In the RL stage, we further optimize the model on a 10K subset of Thyme-RL. All training stages are conducted for one epoch. The maximum latent grid size is set to $M_H = M_W = 16$, yielding up to 256 tokens per latent block. Additional implementation details are provided in the supplementary material.

\noindent\textbf{Evaluated Benchmarks.} 
We evaluate DeepLatent on 13 benchmarks spanning three categories.
\emph{(1) High-Resolution Perception} includes V*~\cite{vstar}, HRBench4K, HRBench8K~\cite{hrbench}, MME-RealWorld, MME-RealWorld-Lite~\cite{mme_real}, and VisualProbe-Hard~\cite{mini_o3}. 
\emph{(2) General Evaluation} includes OCRBench~\cite{ocrbench}, ChartQA~\cite{chartqa}, MMStar~\cite{mmstar}, and RealWorldQA~\cite{rwqa}. 
\emph{(3) Visual Reasoning} includes VisuLogic~\cite{visulogic}, TreeBench~\cite{treebench}, and BabyVision~\cite{babyvision}. We adopt the VLMEvalKit framework for fair evaluation.

\subsection{Evaluation Results}

We report results for both the intermediate SFT model and the final RL models, denoted as DeepLatent-SFT, DeepLatent-RL, and DeepLatent-RL* in Tab.~\ref{tab:evaluation_results}. For brevity, we refer to the final RL models as DeepLatent and DeepLatent* below.

\noindent\textbf{High-Resolution Perception.}
DeepLatent delivers strong gains on high-resolution perception. It achieves 84.8\% on V*, 77.3\% on HR4K, and 74.0\% on HR8K, improving over Qwen2.5-VL-7B by 8.3\%, 8.5\%, and 8.7\%, respectively. It also consistently outperforms prior latent reasoning methods and reaches or exceeds tool-augmented methods, surpassing Thyme by 0.3\% on HR4K and 2.0\% on HR8K, and DeepEyes by 3.6\% on VisualProbe-Hard. On MME-RealWorld and MME-RealWorld-Lite, DeepLatent achieves 64.2\% and 53.7\%, improving over the base model by 6.9\% and 9.4\%. The DeepLatent* variant trained with visual search data further improves the results to 85.3\% on V*, 77.9\% on HR4K, and 74.1\% on HR8K.

\noindent\textbf{General Multimodal Understanding.}
While substantially improving visual perception and reasoning, DeepLatent also preserves strong general multimodal understanding. It achieves 86.4\% on OCRBench and 86.4\% on ChartQA, comparable to the base model, and further obtains 65.0\% on MMStar and 71.5\% on RWQA, improving over Qwen2.5-VL-7B by 3.1\% and 4.1\%, respectively.

\noindent\textbf{Visual Reasoning.}
DeepLatent also brings consistent gains on visual reasoning tasks. It achieves 25.3\% on VisuLogic, 39.0\% on TreeBench, and 16.2\% on BabyVision, improving over Qwen2.5-VL-7B by 5.3\%, 2.0\%, and 3.3\%, respectively. Compared with DeepLatent*, which is trained only with visual search data, DeepLatent achieves more balanced performance and stronger results across all reasoning benchmarks.

We further evaluate DeepLatent on Jigsaw and Multi-view Reasoning from BLINK~\cite{blink}, and Visual Spatial Planning~\cite{vsp}, as shown in Tab.~\ref{tab:visual_reasoning}. DeepLatent achieves 68.0\% on Jigsaw and 59.4\% on Multi-view, outperforming the best compared methods by 16.0\% and 8.3\%, respectively. On VSP, it reaches 83.7\%, surpassing Mirage by 7.7\%, demonstrating stronger spatial path reasoning ability.

{
\setlength{\intextsep}{6pt}
\begin{table}[h]
    \centering
    \small
    \setlength{\abovecaptionskip}{5pt}
    \setlength{\belowcaptionskip}{-3pt}
    \begin{tabular}{l | c c c}
        \toprule
        Model & Jigsaw      & Multi-view & VSP \\
        \midrule
        
        Qwen2.5-VL-7B       & 52.0      & 45.9      & 13.5        \\
        InternVL3-8B        & 50.0      & 51.1      & -         \\
        \midrule

        Pixel-Reasoner-7B       & 52.7      & -         & -         \\
        LVR-7B                  & 52.0      & 46.6      & -         \\
        Monet-7B                & 50.0      & 47.4      & -         \\
        Mirage-7B               &   -       &  -        & 76.0         \\ 

        \midrule
        DeepLatent-RL-7B        & 68.0      & 59.4      & 83.7 \\
        
        \bottomrule
    \end{tabular}
    \caption{Performance on spatial visual reasoning tasks.}
    \label{tab:visual_reasoning}
\end{table}
}

\noindent\textbf{RL Gains.}
DeepLatent-SFT already provides a strong starting point, surpassing most prior methods, and the RL stage further improves most benchmarks. It brings a 3.7\% gain on VisualProbe-Hard, improves MMStar and RWQA by 1.0\% on average, and raises the three visual reasoning benchmarks by 1.1\% on average. These gains show that reinforcement learning further strengthens DeepLatent after supervised training. Overall, DeepLatent demonstrates that the latent reasoning paradigm can reach the effectiveness of prior tool-augmented methods.

\noindent\textbf{Visual Thought Efficiency.}
We compare visual thought generation efficiency in Tab.~\ref{tab:latency} under the same hardware and configuration. DeepLatent achieves a TPLT of 1.0ms and a total latent generation time of 86.1ms, 35$\times$ faster per latent token and 2.6$\times$ faster in total than Monet. This comes from generating all latent tokens in parallel within a single forward pass, even though DeepLatent produces 12.7$\times$ as many latent tokens per sample. Compared with tool-based methods, DeepLatent is over 27$\times$ faster than Thyme (2375ms) and over 38$\times$ faster than DeepEyes (3324ms).

{
\setlength{\intextsep}{6pt}
\begin{table}[h]
\centering
\setlength{\abovecaptionskip}{5pt}
\setlength{\belowcaptionskip}{-7pt}
\small
\setlength{\tabcolsep}{2pt}

\begin{tabular}{l | c c c}
\toprule
Model & TPLT & Latent Tokens & Thought Time \\
\midrule
Thyme-7B         & -      & -     & 2375ms \\
DeepEyes-7B      & -      & -     & 3324ms \\
Monet-7B         & 34.9ms & 6.5   & 227ms  \\
\midrule
DeepLatent-RL-7B & 1.0ms  & 82.7  & 86.1ms \\
\bottomrule
\end{tabular}
\caption{Visual thought generation efficiency on 200 samples. 
TPLT denotes time per latent token. 
Latent Tokens denotes average tokens per latent block. 
Thought Time measures one visual thought step: a tool call or complete latent block generation time.}
\label{tab:latency}
\end{table}
}

\subsection{Ablation Study}

\noindent\textbf{How Visual Latent Helps.}
We compare four SFT methods in Tab.~\ref{tab:ablation_cot}: \textit{Text-only CoT} removes intermediate images and keeps textual reasoning; \textit{Visual-only CoT} keeps intermediate images and answers only; \textit{Interleaved CoT (w/o distill)} uses interleaved data without latent distillation loss; and \textit{Interleaved CoT} adds latent distillation. All variants improve high-resolution perception over Qwen2.5-VL-7B, while \textit{Interleaved CoT} achieves the best results across all metrics, improving V* from 76.5\% to 84.3\% and HR4K from 68.8\% to 76.4\%. \textit{Text-only CoT} consistently outperforms \textit{Visual-only CoT}, showing that visual latents work best when coupled with textual CoT rather than used alone. Removing distillation also degrades performance, confirming the importance of distillation supervision.

{
\setlength{\intextsep}{6pt}
\begin{table}[h]
    \centering
    \small
    \setlength{\tabcolsep}{2pt} 
    \setlength{\abovecaptionskip}{5pt}
    \setlength{\belowcaptionskip}{-5pt}
    \begin{tabular}{l | c c c c}
        \toprule
        Configuration & V* & HR4K  & MME-RW-Lite & MMStar \\
        \midrule
        Qwen2.5-VL-7B              &76.5       &68.8       &44.3  &61.9     \\
        \midrule
        + Text-only CoT            &81.3       &72.3       &50.2  &63.2   \\
        + Visual-only CoT          &80.7       &71.1       &47.5  &57.5     \\
        + Interleaved CoT$^\dagger$& 82.1       & 73.4      & 51.3     & 63.3    \\
        + Interleaved CoT          &84.3       &76.4       &52.4  &63.6  \\
        \bottomrule
    \end{tabular}
    \caption{Ablation study of CoT modality for DeepLatent-SFT. $^\dagger$: without latent distillation loss.}
    \label{tab:ablation_cot}
\end{table}
}

\noindent\textbf{Parallel vs.\ Autoregressive Latent.}
Fig.~\ref{fig:ablation_latent_size} compares two latent generation paradigms and two scaling strategies. \textit{Dynamic} predicts the latent size $(H_z, W_z)$, and \textit{Fixed} always uses a preset maximum size. Parallel generation improves with more latent tokens and nearly saturates at 256 tokens. 
Autoregressive generation peaks near 16 tokens, but additional tokens rapidly degrade performance, whereas parallel generation remains stable even with 576 tokens. This supports our proposed parallel latent generation. For scale allocation, \textit{Fixed} is slightly better at small budgets, while \textit{Dynamic} performs better at larger sizes by preserving the assistant image aspect ratio.

{
\setlength{\intextsep}{4pt}
\begin{figure}[h]
    \centering
    \setlength{\abovecaptionskip}{3pt}
    \setlength{\belowcaptionskip}{-3pt}
    \includegraphics[width=\linewidth]{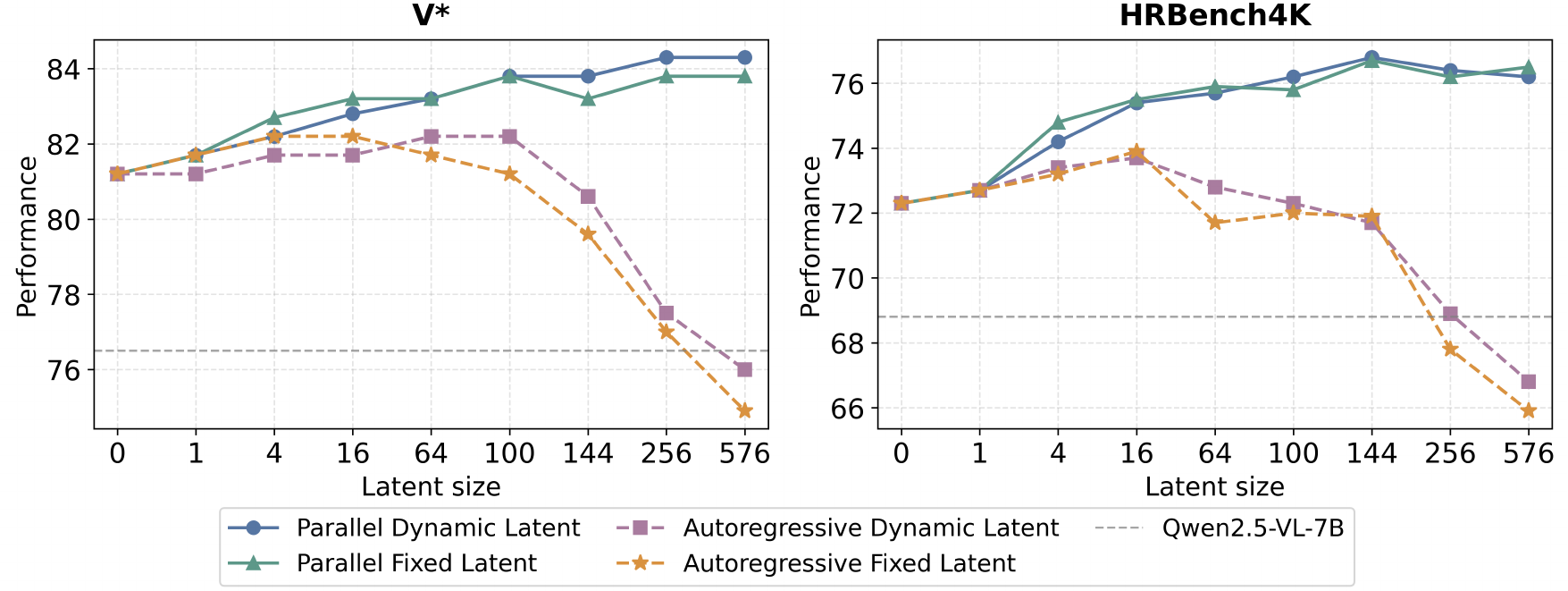}
    \caption{Parallel vs.\ autoregressive generation and dynamic vs.\ fixed scale under varying latent sizes.}
    \label{fig:ablation_latent_size}
\end{figure}
}

Fig.~\ref{fig:ablation_latent_attention_vis} shows latent-to-image attention visualization at layers 8, 16, and 24 of LLM. DeepLatent consistently focuses on the queried ATTENTION sign across layers, while Monet spreads attention over background regions. This suggests that image-grounded parallel latents preserve clearer spatial evidence for fine-grained visual reasoning.

{
\setlength{\intextsep}{4pt}
\begin{figure}[H]
    \centering
    \setlength{\abovecaptionskip}{2pt}
    \setlength{\belowcaptionskip}{-11pt}
    \includegraphics[width=\linewidth]{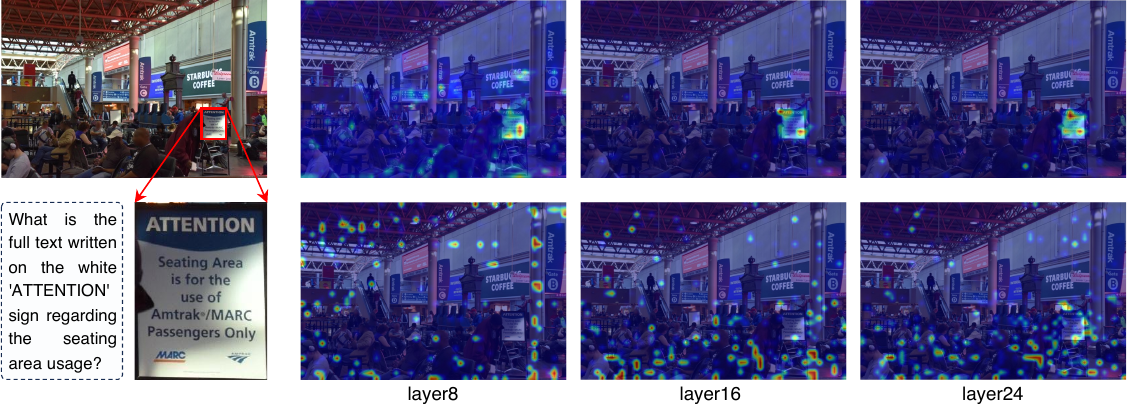}
    \caption{Latent-to-image attention visualization across LLM layers. Top: DeepLatent. Bottom: Monet.}
    \label{fig:ablation_latent_attention_vis}
\end{figure}
}

\noindent\textbf{LatentFormer Parameters.}
We examine the LatentFormer depth and modulation parameters on DeepLatent-SFT, as shown in Tab.~\ref{tab:ablation_latent_former}. Removing LatentFormer feeds learnable latents directly into the LLM without directly extracting image features, causing drops on HR4K (-1.6\%) and VisuLogic (-2.1\%). One layer brings most of the performance improvement, while adding a second layer provides a slight gain. Between the two modulation parameters, latent modulation is more critical, causing an average drop of 1.4\% when removed, while removing image modulation leads to 0.7\% degradation.

{
\setlength{\intextsep}{6pt}
\begin{table}[h]
    \centering
    \small
    \setlength{\tabcolsep}{2pt} 
    \setlength{\abovecaptionskip}{5pt}
    \setlength{\belowcaptionskip}{-5pt}
    \begin{tabular}{l | c c c c}
        \toprule
        Configuration & V* & HR4K & MMStar & VisuLogic \\
        \midrule
        0 (w/o LatentFormer)    & 83.2       & 74.8       & 62.8        & 22.7      \\
        1 layer (default)       & 84.3       & 76.4       & 63.6        & 24.8      \\
        2 layers                & 84.3       & 76.8       & 63.7        & 25.0      \\
        \midrule
        w/o latent modulation   & 82.7       & 75.1       & 62.9        & 22.9      \\
        w/o image modulation    & 83.2       & 76.0       & 63.4        & 23.6      \\
        \bottomrule
    \end{tabular}
    \caption{Ablation study of LatentFormer parameters.}
    \label{tab:ablation_latent_former}
\end{table}
}

\noindent\textbf{Continuous-Space GRPO.}
We compare RL optimization strategies in Tab.~\ref{tab:ablation_rl}. \textit{Text GRPO} applies GRPO only to text and \texttt{<latent>} tokens, while \textit{Joint GRPO} additionally optimizes continuous latent modulation. Text GRPO brings no consistent improvement, and updating LatentFormer further degrades performance by 1.0\% on average, suggesting that learning latent construction with standard GRPO may disrupt learned latent visual semantics. In contrast, Joint GRPO improves over DeepLatent-SFT by 0.7\% on average, with the largest gain on MMStar (+1.4\%). Freezing LatentFormer during Joint GRPO yields more stable gains.

{
\setlength{\intextsep}{6pt}
\begin{table}[h]
    \centering
    \small
    \setlength{\tabcolsep}{3pt}
    \setlength{\abovecaptionskip}{5pt}
    \setlength{\belowcaptionskip}{-9pt}
    \begin{tabular}{l | c c c c}
        \toprule
        Configuration           & V*        & HR4K          & MMStar    & VisuLogic \\
        \midrule
        DeepLatent-SFT          & 84.3      & 76.4          & 63.6      & 24.8    \\
        + Text GRPO             & 83.8      & 76.5          & 63.4      & 24.6    \\
        + Text GRPO$^\dagger$   & 83.2      & 75.3          & 63.1      & 23.6   \\
        + Joint GRPO            & 84.8      & 76.8          & 65.0      & 25.3    \\
        + Joint GRPO$^\dagger$  & 84.3      & 76.1          & 64.7      & 24.9    \\
        \bottomrule
    \end{tabular}
    \caption{Ablation experiments on RL optimization strategies. $^\dagger$: LatentFormer is not frozen in RL stage.}
    \label{tab:ablation_rl}
\end{table}
}

\section{Conclusion}

We presented DeepLatent, a parallel latent visual reasoning framework enabling VLMs to think with images in latent space. By generating image-grounded latents through LatentFormer and optimizing continuous modulation with reinforcement learning, DeepLatent improves visual perception and reasoning while preserving general multimodal understanding. We introduced DeepLatent-180K for latent visual reasoning. Experiments show that DeepLatent outperforms prior latent methods, reaches or exceeds tool-augmented models, and provides more efficient visual thought generation.

\bibliography{refs}

\appendix

\section{Appendix}

\subsection{Implementation Details}
\label{sec:implementation_details}

This section summarizes the main implementation and training hyperparameters used in our supervised fine-tuning (SFT) and reinforcement learning (RL) stages. We use Qwen2.5-VL-7B as the base model and train all models on 8$\times$ NVIDIA A800 GPUs. The SFT pipeline contains two stages: Stage~1 trains on interleaved image-text trajectories with explicit assistant images, and Stage~2 replaces these assistant images with latent blocks for latent distillation. Both SFT stages are trained with the \texttt{trl} framework, while RL training is conducted with the \texttt{verl} framework. We freeze the visual encoder in all training pipelines.

\begin{table}[h]
\centering
\small
\begin{tabular}{l c}
\toprule
Hyperparameter & Value \\
\midrule
learning rate & 0.00001 \\
batch size & 1 \\
gradient accumulation steps & 16 \\
weight decay & 0.01 \\
SFT Stage~1 steps & 1 epoch \\
SFT Stage~2 steps & 1 epoch \\
Max latent size & $16\times16$ \\
Max latent blocks & 5 \\
Stage~1 max question image pixels & $6000\times28\times28$ \\
Stage~2 max question image pixels & $2500\times28\times28$ \\
Stage~2 alignment loss weight & 4.0 \\
Stage~2 scale loss weight & 0.5 \\
\bottomrule
\end{tabular}
\caption{Hyperparameters for SFT.}
\label{tab:sft_hyperparameters}
\end{table}

\begin{table}[h]
\centering
\small
\begin{tabular}{l c}
\toprule
Hyperparameter & Value \\
\midrule
learning rate & 0.000001 \\
batch size & 64 \\
weight decay & 0.01 \\
rollout size & 8 \\
temperature & 0.5 \\
max response length & 2048 \\
training samples & 10,000 \\
training steps & 1 epoch \\
Max latent size & $16\times16$ \\
Max latent blocks & 5 \\
max question image pixels & $2500\times28\times28$ \\
accuracy threshold & 0.6 \\
\bottomrule
\end{tabular}
\caption{Hyperparameters for RL.}
\label{tab:rl_hyperparameters}
\end{table}

\begin{table}[h]
\centering
\small
\begin{tabular}{l c}
\toprule
Hyperparameter & Value \\
\midrule
temperature & 0.5 \\
top\_p & 0.5 \\
top\_k & 5 \\
max\_new\_tokens & 2048 \\
max\_image\_pixels & $16384\times28\times28$ \\
\bottomrule
\end{tabular}
\caption{VLMEvalKit evaluation configuration.}
\label{tab:vlmevalkit_config}
\end{table}

\subsection{Additional RL Details}
\label{sec:appendix_rl_details}

\paragraph{Reward design.}
We use a weighted reward composed of an accuracy reward and a format reward. The accuracy reward is 1 if the final answer is correct and 0 otherwise. In our implementation, correctness is judged by a large VLM. The format reward encourages the model to use the \texttt{boxed\{\}} answer format. The final reward is computed as:
\begin{equation}
R = 0.9 \cdot R_{\mathrm{Acc}} + 0.1 \cdot R_{\mathrm{Format}},
\end{equation}
where $R_{\mathrm{Acc}} \in \{0,1\}$ denotes the accuracy reward and $R_{\mathrm{Format}} \in \{0,1\}$ denotes the format reward.

We additionally test a latent-invocation reward that encourages active use of the latent interface. If a sampled response emits the \texttt{<latent>} token (a regular token in the text vocabulary), this reward is 1; otherwise, it is 0. This term is only used in the ablation below and is not included in the final reward in Eq.~(1). We compare trigger counts with and without this term in Fig.~\ref{fig:latent_trigger_stats}.

{
\setlength{\intextsep}{6pt}
\begin{figure}[h]
    \centering
    \setlength{\abovecaptionskip}{4pt}
    \setlength{\belowcaptionskip}{-8pt}
    \subfloat[With latent invocation reward.]{
        \includegraphics[width=\linewidth]{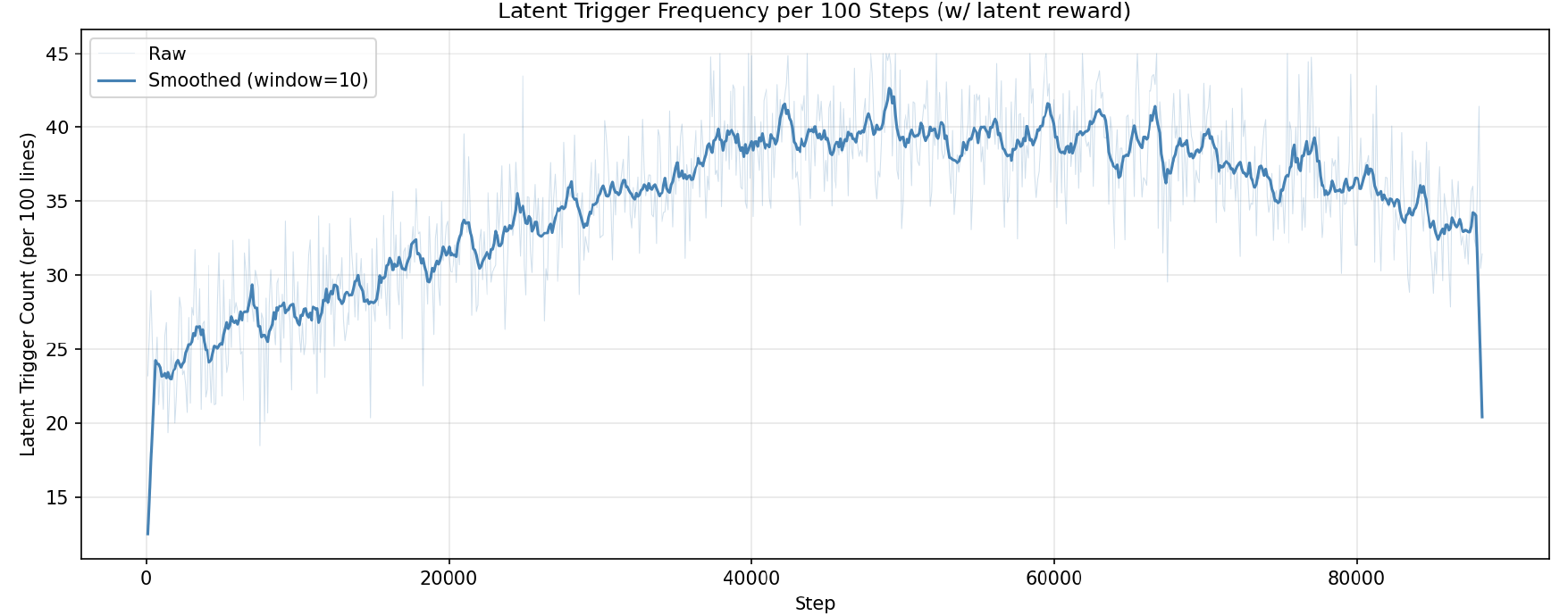}
        \label{fig:latent_trigger_with_reward}
    }
    \\
    \subfloat[Without latent invocation reward.]{
        \includegraphics[width=\linewidth]{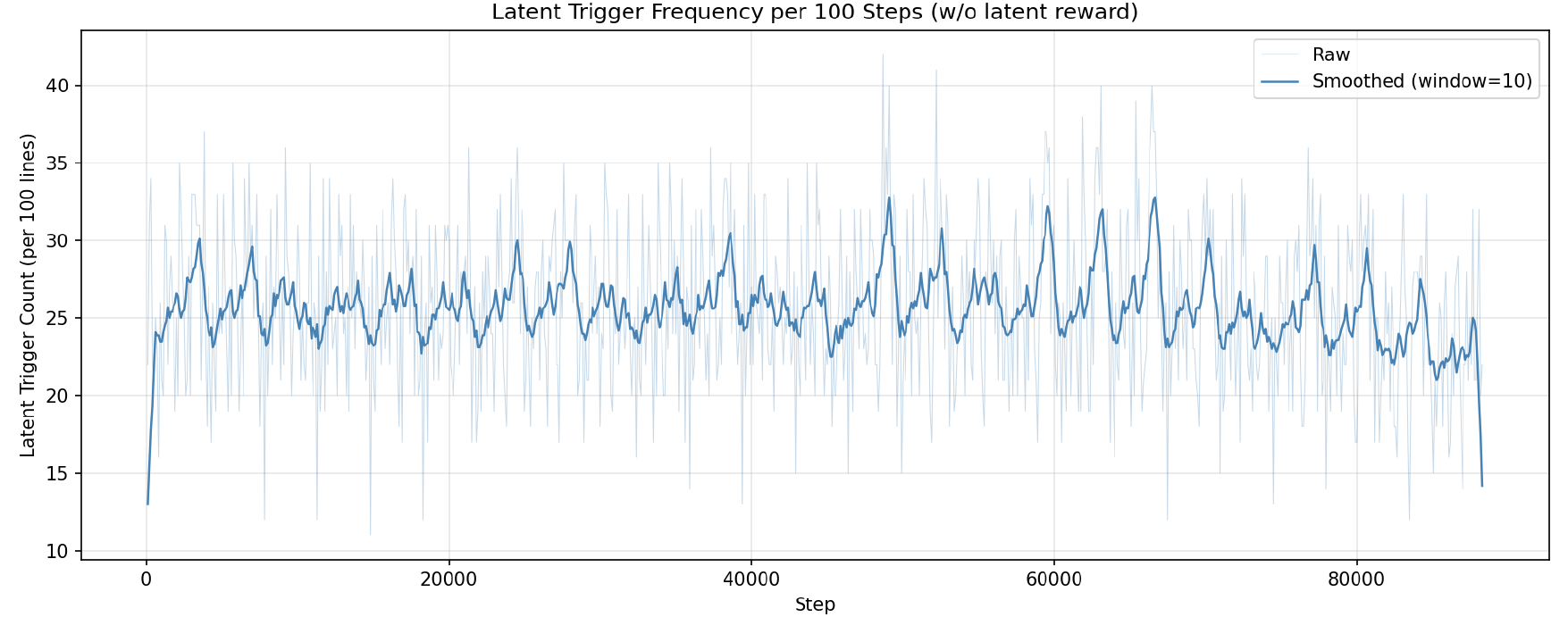}
        \label{fig:latent_trigger_without_reward}
    }
    \caption{Number of active latent invocations during RL training with and without the latent invocation reward.}
    \label{fig:latent_trigger_stats}
\end{figure}
}

The latent invocation reward clearly increases the model's tendency to use latent reasoning during RL. With the reward, the smoothed trigger count rises from about 25 to around 40 per 100 samples and remains at a high level for most of training. Without the reward, the trigger count stays mostly around 25 and shows no clear upward trend. In both settings, the trigger count shows a mild decreasing trend near the end of training. Although the latent invocation reward encourages more frequent latent usage, it does not lead to actual performance improvement. Instead, it often causes the model to generate multiple meaningless latent blocks within a single sampled response. Therefore, we do not adopt the latent invocation reward in the final RL setting.

\paragraph{Continuous-action ratio and KL.}
For continuous modulation actions, we compute policy ratios with diagonal Gaussian densities. Let $\mathbf{a}^{\mathrm{old}}$ denote a rollout action (e.g., $\tilde{\mathbf{L}}^{\mathrm{old}}$ or $\tilde{\mathbf{I}}^{\mathrm{old}}$). The current and old policies define:
\begin{equation}
\begin{aligned}
\pi_{\theta}(\mathbf{a})
&= \mathcal{N}(\mathbf{a}; \boldsymbol{\mu}_{\theta}, \boldsymbol{\sigma}_{\theta}^{2}), \\
\pi_{\mathrm{old}}(\mathbf{a})
&= \mathcal{N}(\mathbf{a}; \boldsymbol{\mu}_{\mathrm{old}}, \boldsymbol{\sigma}_{\mathrm{old}}^{2}).
\end{aligned}
\end{equation}
The ratio for each sampled action is:
\begin{equation}
r_t(\theta)=\exp\left(\log \pi_{\theta}(\mathbf{a}^{\mathrm{old}})-\log \pi_{\mathrm{old}}(\mathbf{a}^{\mathrm{old}})\right),
\end{equation}
with
\begin{equation}
\log \pi(\mathbf{a})=\sum_j \log \mathcal{N}(a_j;\mu_j,\sigma_j^2).
\end{equation}
Here, the action is the modulated output, while the Gaussian mean is parameterized by base variables. For latent-query modulation, $\boldsymbol{\mu}_{\theta}=\boldsymbol{\gamma}_{\theta,l}\odot\mathbf{L}+\boldsymbol{\beta}_{\theta,l}$; for image-feature modulation, $\boldsymbol{\mu}_{\theta}=\boldsymbol{\gamma}_{\theta,i}\odot\mathbf{I}_0+\boldsymbol{\beta}_{\theta,i}$, following Eq.~\eqref{eq:modulation}.

The KL term between the current policy and the SFT reference policy is computed in closed form:
\begin{equation}
\begin{aligned}
\mathbb{D}_{\mathrm{KL}}(\pi_{\theta} \| \pi_{\mathrm{ref}})
&= \frac{1}{2}\sum_j \Bigg[
\log \frac{\sigma_{\mathrm{ref},j}^{2}}{\sigma_{\theta,j}^{2}}
+ \frac{\sigma_{\theta,j}^{2}}{\sigma_{\mathrm{ref},j}^{2}} \\
&\quad + \frac{(\mu_{\theta,j} - \mu_{\mathrm{ref},j})^2}{\sigma_{\mathrm{ref},j}^{2}} - 1
\Bigg].
\end{aligned}
\end{equation}
We compute this KL per continuous action position and then average over positions and samples. For discrete text tokens (including \texttt{<latent>}, which is in the shared vocabulary), we use the standard categorical ratio and KL.

\end{document}